\documentclass{article}
\usepackage{ijcai25}

\usepackage{amsmath,amsfonts,bm}

\def\eqref#1{equation~\ref{#1}}

\def\1{\bm{1}}

\def\vmu{{\bm{\mu}}}
\def\vtheta{{\bm{\theta}}}

\def\vb{{\bm{b}}}

\def\vx{{\bm{x}}}
\def\vy{{\bm{y}}}

\def\mI{{\bm{I}}}

\def\mU{{\bm{U}}}
\def\mV{{\bm{V}}}
\def\mW{{\bm{W}}}
\def\mX{{\bm{X}}}

\DeclareMathAlphabet{\mathsfit}{\encodingdefault}{\sfdefault}{m}{sl}
\SetMathAlphabet{\mathsfit}{bold}{\encodingdefault}{\sfdefault}{bx}{n}

\usepackage{times}
\usepackage{soul}
\usepackage{url}
\usepackage[hidelinks]{hyperref}
\usepackage[utf8]{inputenc}
\usepackage[small]{caption}
\usepackage{graphicx}
\usepackage{amsmath}
\usepackage{amsthm}
\usepackage{booktabs}
\usepackage{algorithm}
\usepackage{algorithmic}
\usepackage[switch]{lineno}

\usepackage{natbib}
\usepackage{cleveref}
\usepackage{subcaption}
\usepackage{multirow}

\title{Identifying Drivers of Predictive Aleatoric Uncertainty}

\author{
Pascal Iversen$^{1,2,}$\thanks{Equal contribution.}\and
Simon Witzke$^{1,}$\footnotemark[1]\and
Katharina Baum$^{1,2,3,}$\thanks{Shared senior authorship.}\And
Bernhard Y. Renard$^{1,2,3,}$\footnotemark[2]\\
\affiliations
$^1$Hasso Plattner Institute, Digital Engineering Faculty, University of Potsdam, Germany\\
$^2$Freie Universität Berlin, Department of Mathematics and Computer Science, Berlin, Germany\\
$^3$Windreich Department of Artificial Intelligence and Human Health \& Hasso Plattner Institute
for Digital Health at Mount Sinai, Icahn School of Medicine at Mount Sinai, New York, USA\\
\emails
katharina.baum@fu-berlin.de,
bernhard.renard@hpi.de 
}

\begin{document}

\maketitle

\begin{abstract}
Explainability and uncertainty quantification are key to trustable artificial intelligence. However, the reasoning behind uncertainty estimates is generally left unexplained. Identifying the drivers of uncertainty complements explanations of point predictions in recognizing model limitations and enhancing transparent decision-making. So far, explanations of uncertainties have been rarely studied. The few exceptions rely on Bayesian neural networks or technically intricate approaches, such as auxiliary generative models, thereby hindering their broad adoption. We propose a straightforward approach to explain predictive aleatoric uncertainties. We estimate uncertainty in regression as predictive variance by adapting a neural network with a Gaussian output distribution. Subsequently, we apply out-of-the-box explainers to the model's variance output. This approach can explain uncertainty influences more reliably than complex published approaches, which we demonstrate in a synthetic setting with a known data-generating process. We substantiate our findings with a nuanced, quantitative benchmark including synthetic and real, tabular and image datasets. For this, we adapt metrics from conventional XAI research to uncertainty explanations. Overall, the proposed method explains uncertainty estimates with little modifications to the model architecture and outperforms more intricate methods in most settings. 
\end{abstract}

\section{Introduction}
Uncertainty quantification and explainability are crucial for adopting machine learning (ML) systems in safety-critical applications, ensuring trust, reliability, and fairness \citep{abdar_review_2021, vilone_explainable_2020, lotsch_explainable_2022}.
Predictive uncertainty in ML refers to the degree of confidence associated with a model's predictions \citep{chua_tackling_2023}. It can be decomposed into an epistemic and aleatoric component \citep{kendall_what_2017}. Epistemic uncertainty stems from data scarcity, such as underrepresented conditions, covariate shift, and model misspecification and can generally be reduced with more data. Aleatoric uncertainty, arising from the random error of the true relationship between inputs, and targets. It reflects irreducible variability in the data. Uncertainty estimation is critical in risk management. It allows taking conservative action, relying on the model only when it exhibits high confidence in its predictions~\citep{kompa_second_2021}. 

Explainability encompasses methods that enhance the transparency of ML models by highlighting how features influence model output or by rendering the internal computations of black-box models more interpretable. Explainability methods enable understanding whether a model has learned relevant patterns from the input data and can reveal interesting, previously unknown associations \citep{samek_explaining_2021, schwalbe_comprehensive_2023}. Uncertainty quantification and explainability ensure accountable, informed, and responsible decision-making and help mitigate biases and risks  \citep{bhatt_uncertainty_2021, mcgrath_when_2023}.

In most applications, explainability focuses on interpreting point predictions~\citep{vilone_explainable_2020}. There is a significant gap in understanding and explaining the drivers of uncertainty estimates. When an ML algorithm is deployed and yields a substantial uncertainty estimate for a specific instance, the possible courses of action involve abstaining from employing the model if alternatives are available or accepting the increased risk. With explainable uncertainties, users gain the capability to identify the factors contributing to elevated uncertainty levels. This understanding allows domain experts to judge their relevance in a given scenario. Additionally, it provides valuable insights into modifications required to augment the model's predictive certainty and performance. In cases where abstaining from model usage is still necessary, factors influencing the decision can be understood and communicated. For example, if such an uncertainty factor is a feature indicating a person's age, it could point to an issue where the model's predictions are more uncertain for specific age groups, even if the age distribution is balanced in the training data. This effect would be undetectable by naive explanations. While detecting and explaining distribution shifts and epistemic uncertainty is an equally interesting problem \citep{brown_using_2022},  we focus our work on aleatoric uncertainty. Aleatoric uncertainty estimates and their explanations are relevant for domains where the noise of the outcome of interest is not constant across independent variables, i.e., heteroscedastic settings. In these cases, aleatoric uncertainty explanations offer complementary information to explanations of point predictions, as the relevant variables influencing mean and variance might differ significantly. Heteroscedastic settings emerge, for example, in the estimation of biophysical variables \citep{lazaro-gredilla_retrieval_2014}, the estimation of cosmological redshifts \citep{almosallam_gpz_2016}, and robotics and vehicle control \citep{bauza_probabilistic_2017, smith_heteroscedastic_2018, liu_large-scale_2021}.

Explanations can be categorized as either local (instance-based) or global (across the whole input space) \citep{schwalbe_comprehensive_2023, adadi_peeking_2018}. A local explanation of the model's uncertainty could foster more transparent discussions about ML-assisted decisions and risks, increasing trust. Global explanations serve to detect general drivers of uncertainty and certainty. These can then be leveraged to formulate hypotheses to improve the model or to detect unintended shortcuts in the uncertainty estimation process, such as spurious correlations or biases.
 
There is little prior work on explaining uncertainties, and existing literature mainly focuses on classification tasks and generally relies on Bayesian neural networks (BNNs) or technical intricacies such as auxiliary generative models \citep{antoran_getting_2021, perez_attribution_2022, ley_diverse_2022, wang_gradient-based_2023}. BNNs assign probability distributions to network weights to capture uncertainty \citep{mackay_practical_1992}. However, due to their computational complexity and involved training process, BNNs have not been as widely adopted as classical neural networks \citep{lakshminarayanan_simple_2017}.

\begin{figure}[ht]
    \centering
    \includegraphics[width=0.475\textwidth]{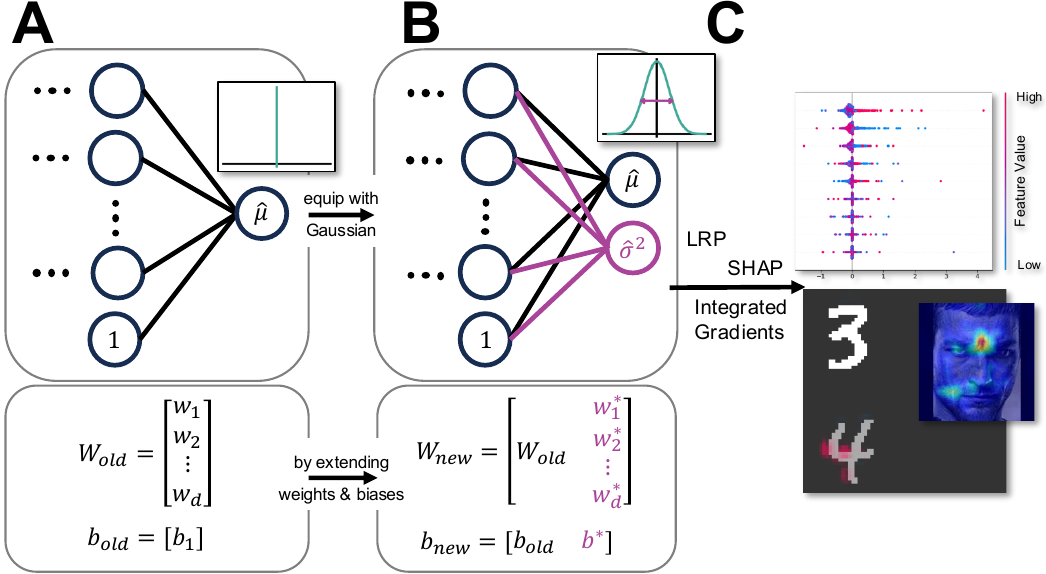}
    \caption{Overview of the variance feature attribution pipeline. (A) A point prediction model with an output layer with weight matrix $\mW_{old} \in \mathbb{R}^{d\times1}$ and a scalar bias. We equip this model with a Gaussian distribution resulting in (B), a model with output weight matrix $\mW_{new} \in \mathbb{R}^{d\times2}$ and bias $\vb_{new} \in \mathbb{R}^2$. The two outputs are the mean $\hat{\mu}$ and the variance $\hat{\sigma}^2$ of the predictive distribution. (C) From there, we can explain the variance using any suitable explainability method, resulting in attributions to the input features that can be used to understand the drivers of the model's aleatoric uncertainty.} \label{fig:fig1}
\end{figure}

We propose a straightforward and scalable approach for explaining uncertainties in a heteroscedastic regression setting that can be readily integrated into ML pipelines (see~\Cref{fig:fig1}). We extend point prediction models to additionally estimate parameters of the spread of a given probability distribution. Specifically, we predict parameters of a Gaussian distribution as in a heteroscedastic regression model \citep{bishop_mixture_1994}. The variance parameter of the Gaussian can be interpreted as a measure of the aleatoric uncertainty of the model. We can then use any explainability method to explain the variance estimate provided by this distributional model. By highlighting input features contributing to the variance output, we identify the inputs contributing to model uncertainty.

Currently, there is a gap in the comparative evaluation of uncertainty explainers in the literature. Therefore, we introduce a benchmark with synthetic data with a known data-generating process to analyze a method's ability to detect uncertainty drivers. In addition, we introduce MNIST+U, an image dataset including known uncertainty drivers based on MNIST \citep{deng_mnist_2012}. We compare our approach to Counterfactual Latent Uncertainty Explanations (CLUE) \citep{antoran_getting_2021} and InfoSHAP \citep{watson_explaining_2023}. For this purpose, we adapt unsupervised XAI metrics to evaluate the uncertainty explainers.

In summary, our contribution is as follows:
We propose a straightforward explanation method for uncertainty and evaluate it against existing approaches. Further, we devise tabular and image benchmarks, including established metrics from the XAI field. 
Thereby, we provide a resource for informed usage of uncertainty explanation methods.

\subsection{Related Work}
In some research communities, such as causal inference, graphical models, and Gaussian processes, explicitly modeling uncertainty is a prominent area of interest. Furthermore, uncertainty quantification and explainability are rich areas of research within the deep learning field \citep{abdar_review_2021, vilone_explainable_2020}.
Yet, few researchers have recognized the importance of explaining the sources of uncertainty in deep learning predictions.
\citet{yang_explainable_2023} have developed an explainable uncertainty quantification approach for predicting molecular properties. They employ message-passing neural networks and generate unique uncertainty distributions for each atom of a molecule. This approach is inherently specialized for graph-based representations of molecules.
CLUE \citep{antoran_getting_2021} and related approaches \citep{perez_attribution_2022, ley_diverse_2022} derive counterfactual explanations by optimizing for an adversarial input that is close to the original input but minimizes uncertainty. The adversarial input is constrained to the data manifold with a deep generative model of the input data to prevent out-of-distribution explanations. This requires an optimization process for each instance's explanation and the training of an auxiliary generative model, rendering CLUE and its extensions computationally demanding and difficult to implement. Additionally, \citet{antoran_getting_2021} developed an evaluation method for contrastive explanations of uncertainty.
\citet{wang_gradient-based_2023} have developed a gradient-based uncertainty attribution method for image classification with BNNs. They modify the backpropagation to attain complete, non-negative pixel attribution.
To detect and explain model deterioration, \citet{mougan_monitoring_2023} use classical ML methods and bootstrapping. They train a model and obtain uncertainty estimates on a test set transformed with an artificial distribution shift. In a second step, they train another model to predict the uncertainty estimates from the first step. Subsequently, Shapley values are estimated for the second model to explain the uncertainty. \citet{mehdiyev_quantifying_2023} employ quantile regression forests to obtain prediction intervals that quantify uncertainty. They extract feature attributions for the uncertainty by estimating Shapley values directly for these prediction intervals as output. 
\citet{watson_explaining_2023} introduce variants of the Shapley value algorithm to explain higher moments of the predictive distribution by quantifying feature contributions to conditional entropy. They use a split conformal inference strategy. They first train a base model to predict conditional probabilities. Subsequently, they fit an auxiliary model to the base model's log square residuals. They interpret the estimated Shapley values of this residual model as uncertainty explanations.
\citet{bley_explaining_2025} propose a second-order uncertainty attribution method that explains predictive uncertainty by computing the covariance of first-order feature attributions across model ensembles. While this approach offers insights into individual and joint feature contributions to uncertainty, it is limited to ensembles and cannot explain aleatoric uncertainty.

\section{Methods}

\subsection{Deep Heteroscedastic Regression and Extension of Pre-trained Models}\label{subsec:dhr}
We explain uncertainties in neural network regressors using deep distributional networks. Specifically, we employ heteroscedastic regression with a Gaussian output to model variance in addition to the mean and, therefore, capture input dependence of the output noise.
Here, we consider a regression setting with $n$ independent training examples $\lbrace (\vx_i, y_i)\rbrace_{i=1}^n$ with input feature vector $\vx_i \in \mathbb{R}^k$ and target $y_i \in \mathbb{R}$, $i=1,\ldots,n$.
Instead of providing a complete picture of the conditional distribution of the target, deep regression models usually only estimate its conditional mean by optimizing the mean squared error (MSE) or comparable loss functions. 
In contrast, we assume a heteroscedastic Gaussian as the conditional distribution $y \mid {\vx} \sim \mathcal{N}\left( \mu_{\vx}, \sigma^2_{\vx} \right)$ and represent its mean $\mu_{\vx}$ and variance $\sigma^2_{\vx}$ using a neural network $f_\theta: \mathbb{R}^k \to \mathbb{R} \times \mathbb{R}^+$ with weights $\vtheta$ and two output neurons producing the mean and variance estimates $f_\vtheta\left({\vx} \right) = (\hat \mu_{\vx}, \hat \sigma^2_{\vx})$,
respectively. As first described by \citet{bishop_mixture_1994}, we can then optimize the Gaussian negative log-likelihood (GNLL):
$
    \mathcal{L} \propto \sum_{i=1}^n \left(\log(\hat \sigma^2_{\vx_i}) + \frac{(y_i-\hat \mu_{\vx_i})^2}{\hat \sigma^2_{\vx_i}} \right)
$
and interpret the predicted variance as a measure of the aleatoric uncertainty of the model.
However, naively optimizing this criterion with overparametrized models such as deep neural networks can be unstable \citep{wong-toi_understanding_2023, nix_estimating_1994, seitzer_pitfalls_2022}. In practice, these convergence difficulties can be mitigated by initially training the model using solely the MSE $\sum_{i=1}^n\left(y_i-\hat \mu_{\vx_i}\right)^2$ and subsequently switching to the GNLL \citep{sluijterman_optimal_2023}.

The two-stage training process aligns with transfer learning: MSE-based initial training serves as pre-training, followed by fine-tuning with the GNLL to capture predictive uncertainty. Extending existing pre-trained models to capture uncertainty is relevant when the model size and associated training costs make full re-training unfeasible. Pre-trained regression models can be extended by concatenating a column of randomly initialized weights to the weight matrix of the output layer to attain a variance estimate (see \Cref{fig:fig1}). 

\subsection{Post-hoc Explanation of Predictive Variance}\label{subsec:uncertaintyXAI}
Classic explainability methods explain the predicted class or point prediction. In contrast, we want to explain the variance output in a heteroscedastic regression model. In these models, variance is an additional output to which we can apply any existing, appropriate explainability method. 
In principle, an uncertainty explanation can be achieved for any parametrized output distribution for which an explicit formulation of the uncertainty is available. In the case of a Gaussian output distribution, the application is most intuitive since its variance parameter is a direct output of the neural network. Furthermore, unlike distributions such as the Poisson or exponential distributions, the variance is uncoupled from the mean output. For binary classification, where the model outputs the parameter of a Bernoulli distribution, an entropy formulation of uncertainty can be utilized. Alternatively, explaining aleatoric uncertainty for classification can be approached by operating in the logit space.

We employ model-agnostic and model-specific post-hoc explainability methods to explain uncertainty. Model-specific methods are limited in the type of models that they can explain but may offer advantages such as lower computational complexity. In contrast, model-agnostic methods can be applied to any model \citep{adadi_peeking_2018}.
For our experiments,  we combine the approach described in \Cref{subsec:dhr} with multiple explainability methods and refer to this conjunction as Variance Feature Attribution (VFA) flavors. As the first explainability method, we use KernelSHAP \citep{lundberg_unified_2017}, a model-agnostic, local explainability method. KernelSHAP approximates Shapley values using a weighted linear surrogate model with an appropriate weighting kernel (VFA-SHAP). Additionally, for image tasks, we leverage DeepSHAP \citep{lundberg_unified_2017}, a variant of SHAP tailored for deep learning models, which combines SHAP values with the DeepLIFT algorithm \citep{shrikumar_learning_2017} to efficiently attribute model predictions to input features. We also employ Integrated Gradients (IG) \citep{sundararajan_axiomatic_2017}, which is a local, model-specific method and assigns feature importance by integrating predictions over a straight path from a baseline to the input (VFA-IG).
Further, we use Layer-Wise Relevance Propagation (LRP) \citep{bach_pixel-wise_2015}, a local, model-specific explainability method developed for neural networks where the importance is distributed backward to the input layer by layer weighted by a neuron's contribution (VFA-LRP). We compare the VFA flavors to CLUE, for which we have to train a variational autoencoder on the train data and apply the optimization as detailed by \citet{antoran_getting_2021}. CLUE attributions are the absolute differences between counterfactual and input feature vectors. CLUE is local and model-specific. Further, we reimplement InfoSHAP for regression, which estimates the uncertainty using an auxiliary model trained on the log-square residuals of a base model. The uncertainty attribution is attained by estimating the Shapley values of the auxiliary model \citep{watson_explaining_2023}. As InfoSHAP builds on SHAP, it is a model-agnostic, local explainability method. 
Global explanations are obtained by averaging local method results over a dataset.

\subsection{Uncertainty Explanation Evaluation Metrics}
There is little prior work on evaluating the quality and properties of uncertainty explanations. Generally, high-quality explanations have to be robust, faithful, and highlight relevant input features. We extend established metrics for general XAI to the explanation of model uncertainty. 

In a situation where ground truth noise drivers are known, we can examine if explanation methods correctly rediscover them.  \citet{arras_clevr-xai_2022} introduce metrics for this setting for classical XAI: \textit{Relevance Rank Accuracy} (RRA) describes the proportion of known relevant features that are rediscovered by the explanation method for a given sample $\vx_i$.

\textit{Relevance Mass Accuracy} (RMA) describes the amount of relevance that is assigned to the ground truth features, normalized by the total amount of relevance.
For uncertainty explanations, we judge if a method discovers features that correlate with the standard deviation of the target's heteroscedastic noise. To scrutinize global explanations, we apply these accuracy metrics to global feature attributions, giving rise to global relevance rank accuracy (GRA) and global relevance mass accuracy (GMA). Global accuracy measures how effectively a model detects general drivers of uncertainty across the entire dataset. In contrast, local accuracy indicates the model's ability to identify uncertainty sources for individual instances and is, therefore, a stricter criterion.
 
\citet{alvarez-melis_robustness_2018} argue that \textit{Robustness} is a key property of explanations, demanding that proximal inputs lead to similar explanations. They propose to evaluate robustness with local Lipschitz continuity:
$
    \hat{L}(\vx_i) = \max_{\vx_j \in \mathcal{N}_{\epsilon}(\vx_i)}\frac{\left \| f(\vx_i)-f(\vx_j) \right \|_2}{\left \| \vx_i-\vx_j \right \|_2}\text{,}
$
where $f$ is the explanation method. For a dataset with only continuous features, the perturbation space $\mathcal{N}_{\epsilon}(\vx_i)$ is a ball with radius $\epsilon$ around sample $\vx_i$.  However, continuous perturbations lack meaning for categorical features. Instead, the perturbation space is defined as the set of data points close to $\vx_i$: $\mathcal{N}_{\epsilon}(\vx_i)=\{\vx_j \in \mathcal X | \left \| \vx_i-\vx_j \right \|\ \leq \epsilon, \vx_i \neq \vx_j\}$, where  $\mathcal X$ is the set of test inputs. Low Lipschitz estimates indicate small changes in the explanation upon perturbation and, therefore, high robustness. This notion of robustness can be extended to uncertainty explanations by applying it to the variance head predictions or an auxiliary uncertainty model.

Further, we analyze \textit{Faithfulness} of the explanations. If an explanation is faithful, changing input features that are considered relevant should lead to a significant reduction in prediction performance. Commonly, this is measured as the increase of the loss upon perturbation of relevant features \citep{arras_clevr-xai_2022}. However, the GNLL loss we use during training is a function of the mean and variance, and its magnitude is not interpretable. We aim to evaluate the perturbation's impact on the quality of the uncertainty estimate.  Naturally, we demand that a higher uncertainty estimate should relate to a higher expected squared error of the mean prediction. Therefore, we measure the correlation between the squared residuals and the uncertainty estimates. Let $\mathbf{y} \in \mathbb{R}^{n}$ be the vector of ground truth target values from the test set, and let $\hat{\bm{\mu}}(\mathbf{X}) \in \mathbb{R}^{n}$ and $\hat{\bm{\sigma}}^2(\mathbf{X}) \in \mathbb{R}^{n}$ denote the predicted means and variances, respectively, for test inputs $\mathbf{X} \in \mathbb{R}^{n \times d}$, where $n$ is the number of test samples and $d$ is the number of features. Let $\mathbf{A} \in \mathbb{R}^{n \times d}$ be the feature attribution matrix, where $\mathbf{A}_{ij}$ is attribution of feature $j$ for sample $i$, as produced by an uncertainty explainer applied to $\hat{\bm{\sigma}}^2(\mathbf{X})$. We first calculate the Spearman correlation $\rho_s =\text{corr}_s\left((\vy-\hat \vmu(\mX))^2, \bm{\hat \sigma}^2(\mX)\right)$.
We then determine the $k$ globally most important uncertainty features.
Precisely, we compute the index set of the most important features
$
\mathcal{I}_k = \text{top-}k_j \left( \frac{1}{n} \sum_{i=1}^n |\mathbf{A}_{ij}| \right),
$
i.e., the indices of the $k$ features with the highest average absolute attribution values. Subsequently, we define a perturbed input matrix $\mathbf{X}' \in \mathbb{R}^{n \times d}$ by adding Gaussian noise to the $k=3$ most important features for uncertainty prediction
$$
\mathbf{X'}_{ij} =
\begin{cases}
\mathbf{X}_{ij} + \delta_{ij}, & \text{if } j \in \mathcal{I}_k \\
\mathbf{X}_{ij}, & \text{otherwise}
\end{cases}, \quad \delta_{ij} \sim \mathcal{N}(0, 1).
$$
Based on this, we calculate the correlation of the original residuals with the perturbed uncertainties $\rho_s' =\text{corr}_s\left((\vy-\hat\vmu(\mX))^2, \bm{\hat\sigma}^2(\mX')\right)$ and expect the variance to be less expressive after the perturbation, i.e., the change  $\rho_s'-\rho_s$ is negative, if the uncertainty explanation is faithful.

\subsection{Benchmark on Tabular Data}
\subsubsection{Synthetic Data Generation}
\label{subsec:syn-pipeline}
Evaluating explainability methods on real-world data is challenging due to the subjective nature of interpreting explanations based on expert prior knowledge. To address this, we employ synthetic data with a known data-generating process. Thereby, we can introduce controlled sources of heteroscedasticity, which we aim to detect. Specifically, we sample a synthetic ground truth using a linear system $\vmu =  \mV \bm{\beta}$ with a design matrix $\mV \in \mathbb{R}^{n \times p}$ with $\mV_{ij} \sim \mathcal{N}(0,1)$, and ground truth coefficients $\bm \beta \in \mathbb R^p$ with $\bm \beta_i\overset{\text{iid}}{\sim}\text{Uniform}([-1,1])$.
We introduce heteroscedastic noise sources with an absolute-value transformed polynomial model for the heteroscedastic noise standard deviation:
$
    \bm{\sigma} =  \mid \phi(\bm{U}) \bm{\gamma} + \bm{\delta}\mid,
$
whereby $\bm{U} \in \mathbb{R}^{n \times  {p^ \prime}}$ is a design matrix with $\bm{U}_{ij} \overset{\text{iid}}{\sim} \mathcal{N}(0,1)$,
\begin{equation*}
    \phi (u_1, u_2,\ldots, u_{p^ \prime}) \to (1, u_1,\ldots, u_{p^ \prime}, u_1^2, u_1u_2,\ldots, u_{p^ \prime}^2)
\end{equation*} is a second degree polynomial feature map, and $\bm \delta \sim \mathcal{N}\left(\bm 0,\sigma^2_\delta \mI \right)$ is the uncertainty model error. The ground truth noise coefficients $\bm{\gamma} \in \mathbb{R}^{\binom{p^ \prime+2}{2}}$ have entries sampled from $\bm \gamma_i \sim \text{Uniform}([-1, -0.5] \cup [0.5, 1])$ to avoid negligible effects. We can then sample the target $\vy \in \mathbb{R}^n$ with
\begin{equation*}
\label{eq:sampling}
    \vy \sim \mathcal{N}\left(\vmu, \alpha \cdot \text{diag}(\bm{\sigma}^2) + \sigma_{\epsilon}^2 \mI \right),
\end{equation*}
where $\alpha \in \mathbb{R}^+$ determines  the overall strength of the heteroscedastic uncertainty and $\sigma_{\epsilon}^2 \in \mathbb{R}^n$ regulates the homoscedastic noise.

For our experiments, we set $\alpha=2.0$, $\sigma_{\epsilon}^2=0.02$, and $\sigma^2_\delta=0.05$ to get non-negligible, feature-dependent noise.
We choose $p=70$ and $p^\prime = 5$ so that the uncertainty sources have to be detected among a larger set of features that do not influence the uncertainty. We sample $n=41,500$ data points and concatenate both design matrices to attain the input $\mX_{(n \times 75)} = \left[\mU_{(n\times 5)}, \mV_{(n \times 70)}\right]$ which we split into 32,000 train, 8,000 validation, and 1,500 test instances.

In reality, we expect noise features to overlap with features influencing the mean. We separate these in the synthetic data to allow for unambiguous assessments in the evaluation. 
However, our implementation accommodates the analysis of mixed scenarios, where a subset of features simultaneously influences the mean and variance.

\subsubsection{Tabular Real World Datasets}
In addition, we incorporate three standard regression benchmark datasets into our evaluation: UCI Wine Quality \citep{cortez_wine_2009}, Ailerons \citep{torgo_inductive_1999}, and LSAT academic performance \citep{wightman_lsac_1998}. These datasets were selected to vary in size and complexity. The Wine Quality dataset, where we use red wines only, includes 11 features for 1,599 samples. Ailerons has 40 features and 13,750 samples, while LSAT, the largest dataset, has 21,790 samples with two continuous and two one-hot encoded features. All datasets are split into 70\% training, 10\% validation, and 20\% testing.

\subsubsection{Tabular Benchmarking Setup}
\label{a:benchmarking_setup}
We divide our tabular benchmark into two stages. First, we qualitatively and quantitatively evaluate the uncertainty explanation methods in a controlled setting on a synthetically generated dataset. Second, we investigate the same methods concerning their local RRA and RMA, faithfulness, and robustness on synthetic and real-world data. 

The first stage of our benchmark aims to detect global drivers of uncertainty in a synthetic setting. We fit a deep neural network of four hidden layers with 64, 64, 64, and 32 units and two outputs for the mean and variance prediction. We train using dropout on the first two layers, Adam optimizer and a batch size of 64. We pre-train using the MSE and fine-tune the model using the GNLL as the loss function, selecting weights with the lowest validation loss. We attain a global feature importance measure as the mean absolute variance feature attributions over all or a specific subset of test instances, which we then analyze using GRA and GMA.

We follow the same model training procedure for the second benchmarking stage, evaluating accuracy, faithfulness, and robustness. Estimating the local RRA and RMA for a given method requires prior knowledge of features affecting the explained quantity, i.e., uncertainty. As this is not available for our selected real-world datasets, we augment them with synthetic noise that we aim to detect, effectively creating a semi-synthetic setting. For the three real-world datasets, we consider two scenarios. We add five noise features to the datasets and heteroscedastic Gaussian noise to the targets with a standard deviation correlating with these features. Since the real-world datasets are small, we first use a simple noise model where the absolute sum of the noise features is the standard deviation of the noise distribution, a setting referred to as 1-S. In a second scenario, 50-C, we use the more complex polynomial noise model described in \Cref{subsec:syn-pipeline}. To provide more data to the model in the complex noise scenario, we replicate each data point in the train sets 50 times before sampling additional uncertainty features and target noise. For the synthetic datasets, we similarly perform experiments with a simple (S) and complex (C) noise model but without adjusting the dataset size.

We evaluate the robustness of the uncertainty explanation methods for each dataset by estimating the local Lipschitz continuity for 200 randomly selected data points from the test set. For each selected point $\vx_i$, we compute a local Lipschitz estimate $\hat{L}(\vx_i)$ by introducing 100 perturbations. For each feature, we sample a perturbation from a uniform distribution centered at the feature value with a range of $ 2\%$ of the range of the feature in the train set\footnote{\scriptsize \mbox{Adapted\hspace{0.1cm}from\hspace{0.1cm}\url{https://github.com/viggotw/Robustness-of-Interpretability-Methods}}}. This is not applicable to LSAT's categorical features. Instead, we resort to the discrete definition of local Lipschitz continuity. Specifically, we compute $\hat{L}(\vx_i)$ for 200 data points sampled from the test set such that their neighborhood $\mathcal{N}_{\epsilon}(\vx_i)$ with $\epsilon=0.2$ contains more than five instances.

To evaluate faithfulness, we apply standard Gaussian noise to perturb the three globally most important uncertainty drivers of the test data. We omit the LSAT dataset as it mainly contains categorical features for which continuous perturbations lack meaning.

We note that we only add synthetic noise to estimate accuracy metrics, i.e., when calculating RRA and RMA. For all other experiments, we use the real-world datasets as is.

\subsection{Benchmark on Image Data: MNIST+U}
To extend our evaluation to a higher-dimensional problem with more realistic feature dependencies, we consider the task of image regression. We introduce the MNIST+U dataset that extends the original MNIST dataset \citep{deng_mnist_2012} with an uncertainty component. We create 500,000 composite images with labels. For each sample, two $28 \times 28$ MNIST digit images are randomly selected and placed into different corners of a $64 \times 64$ canvas. The first digit is white and represents the mean ($\mu_i$) of a target Gaussian distribution. The second gray digit represents its standard deviation ($\sigma_i$). Thus, we sample the label ($y_i$) as: $y_{i} = \mu_i + \epsilon_i, \quad \epsilon_i \sim \mathcal{N}(0, \sigma_i^2).$

We split the generated data into train, validation, and test sets consisting of 70\%, 10\%, and 20\% of the data, respectively. For the image benchmark, we apply a CNN with two parallel encoders where one predicts the mean and the other estimates the variance. Each encoder has two convolutional layers (16 and 32 filters), max-pooling, and fully connected hidden layers with dropout and 128, 64, and 32 nodes. We train the model with MSE for 16 epochs, then switch to GNLL loss until the validation loss converges. We use the Adam optimizer and a batch size of 256. We evaluate uncertainty explainers using RMA and RRA by comparing assigned pixel relevance to the ground truth variance and mean masks. To account for explanations extending beyond the masked digits, each mask is dilated by two pixels.iments and to create the MNIST+U dataset is available online on GitHub\footnote{\scriptsize\url{https://github.com/DILiS-lab/drivers-of-predictive-aleatoric-uncertainty}}. We further make the MNIST+U dataset available separately on Zenodo\footnote{\scriptsize\url{https://doi.org/10.5281/zenodo.15373739}}.
.

\section{Results}
\subsection{Benchmarking the Detection of Uncertainty Drivers using Synthetic Datasets}

\begin{figure}[!t]
    \centering
    \includegraphics[width=0.475\textwidth, trim = 4 0 4 4, clip]{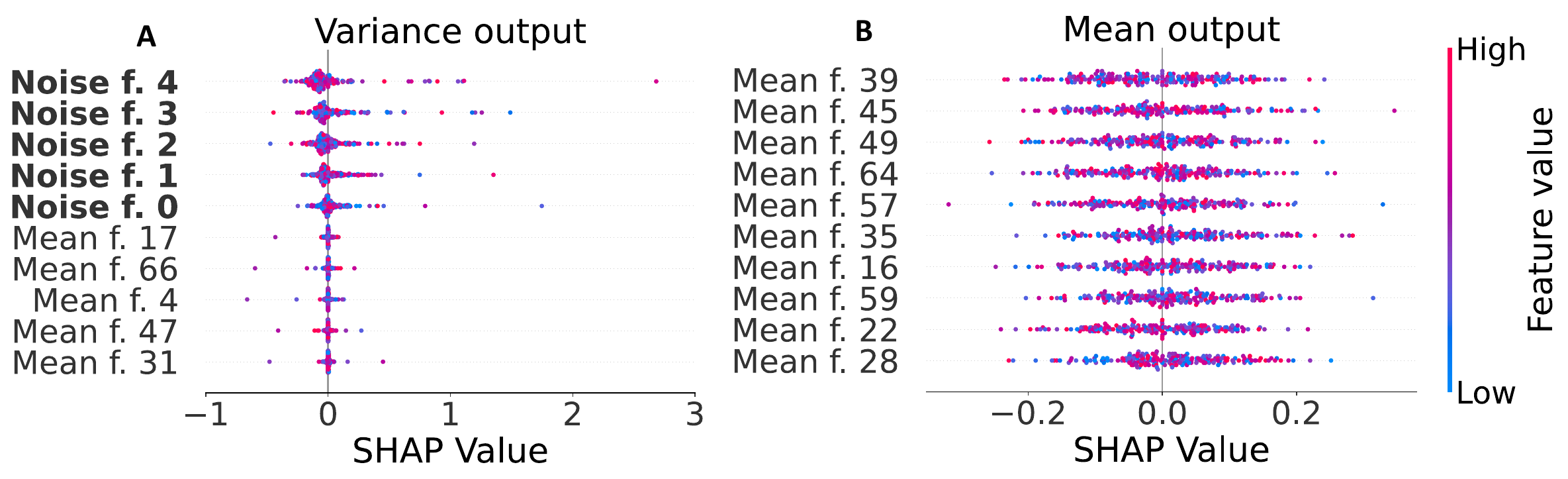}
   
  \caption{Explanations for uncertainty and mean predictions for the synthetic dataset using VFA-SHAP. We display SHAP summaries for the 10 most important features of (A) model uncertainty or (B) mean prediction ordered by the mean of their absolute estimated Shapley values. VFA-SHAP identifies all noise features driving the model's aleatoric uncertainty. Explaining the mean output offers complementary information but does not detect uncertainty features.} 
\label{fig:shap_summary}
\end{figure}

We first examine the capability of VFA-SHAP to identify the drivers of uncertainty, which are features that correlate with the magnitude of the heteroscedastic noise. We know the data-generating process for the synthetic dataset and, therefore, the ground truth noise sources. Using this dataset, VFA-SHAP accurately identifies the five ground-truth noise features driving uncertainty, which are distinct from features influencing the mean (\Cref{fig:shap_summary} A and B). We verify that our model captures uncertainty accurately to ensure meaningful explanations. All trained models are well-calibrated, predictive of model error, and well-suited to our application setting. Details are available in the GitHub repository.

\begin{figure}[!b]
    \centering
    \includegraphics[width=0.48\textwidth]{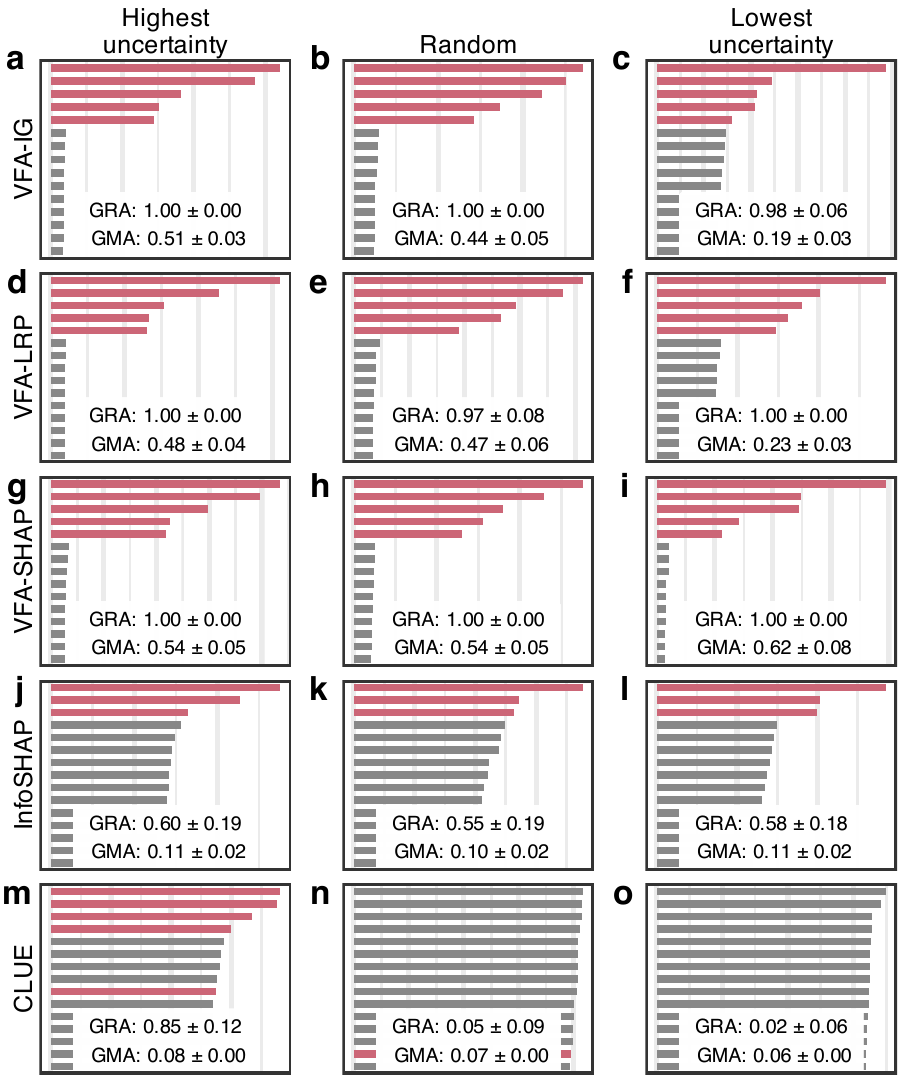}
    \caption{
    Top 15 global importance features with GRA and GMA for each uncertainty explainer. First column: From 1,500 test samples, we explain the 200 instances with the highest predicted uncertainty. VFA flavors highlight the ground truth noise features (red), while Infoshap and CLUE are less accurate. Second and third columns: For 200 random or low uncertainty instances, VFA remains accurate, while CLUE becomes unreliable. InfoSHAP maintains adequate performance but consistently detects only three noise features.}
    \label{fig:cluecomparison}
\end{figure}

Further, we analyze the global uncertainty explanation abilities of all VFA flavors, CLUE and Info\-SHAP (\Cref{fig:cluecomparison}). CLUE is applied to the same neural network as VFA, whereas InfoSHAP utilizes XGBoost.
Uncertainty estimation may facilitate cautious model application only at high certainty or opting out of model usage due to substantial uncertainty. Therefore, in addition to 200 random instances, we apply the explainers to the test set's 200 highest and lowest uncertainty instances. 
We find that VFA flavors and CLUE effectively identify uncertainty drivers for high-uncertainty instances, reflected by their GRAs (with five ground truth noise features) close to $1$. VFA flavors exhibit superior GMA, which signifies their capacity to disregard irrelevant features. VFA performs reliably for random and low uncertainty examples, while CLUE's performance deteriorates. This suggests that, unlike CLUE, VFA can explain the factors contributing to certainty. InfoSHAP, while underperforming for instances with high uncertainty, clearly outperforms CLUE for random and low uncertainty instances. 
We provide code for this figure and further examples showing that VFA considerably outperforms CLUE and InfoSHAP in all three settings.

\subsection{Local Accuracies, Faithfulness, and Robustness}

We evaluate the local RRA and RMA for the real-world datasets and the synthetic dataset in two settings, one simple (1-S, S) and one complex (50-C, C) as described in \Cref{a:benchmarking_setup} (see \Cref{tab:localization}). VFA-SHAP outperforms the other explainers over most datasets. Generally, VFA of any flavor performs best, for the simple settings. However, in the complex setting, InfoSHAP consistently outperforms VFA-IG and VFA-LRP and achieves competitive performance to VFA-SHAP on LSAT and Ailerons. 

\begin{table}[tb]
\centering
\resizebox{0.49\textwidth}{!}{%
\begin{tabular}{llrrrrrrrr}
    \toprule
    \multicolumn{1}{c}{} &
      \multicolumn{1}{c|}{} &
      \multicolumn{2}{c|}{\textbf{Red Wine}} &
      \multicolumn{2}{c|}{\textbf{Ailerons}} &
      \multicolumn{2}{c|}{\textbf{LSAT}} &
      \multicolumn{2}{c}{\textbf{Synthetic}} \\
     &
      \multicolumn{1}{c|}{} &
      1-S &
      \multicolumn{1}{r|}{50-C} &
      1-S &
      \multicolumn{1}{r|}{50-C} &
      1-S &
      \multicolumn{1}{r|}{50-C} &
      S &
      C \\ \midrule
    \multicolumn{10}{l}{\textbf{RRA}} \\ \midrule
        & VFA-IG &
        $0.61 (0.05)$ &
        $0.60 (0.04)$ &
        $0.81 (0.03)$ &
        $0.70 (0.05)$ &
        $0.81 (0.04)$ &
        $0.74 (0.05)$ &
        $0.75 (0.02)$ &
        $0.38 (0.05)$ \\
        
        & VFA-LRP &
        $0.62 (0.05)$ &
        $0.61 (0.04)$ &
        $0.79 (0.04)$ &
        $0.67 (0.06)$ &
        $0.81 (0.02)$ &
        $0.73 (0.06)$ &
        $0.75 (0.01)$ &
        $0.41 (0.05)$ \\
        
        & VFA-SHAP &
        $\textbf{0.85} (0.02)$ &
        $\textbf{0.90} (0.02)$ &
        $\textbf{0.88} (0.01)$ &
        $\textbf{0.88} (0.02)$ &
        $\textbf{0.93} (0.02)$ &
        $\textbf{0.92} (0.02)$ &
        $\textbf{0.85} (0.01)$ &
        $\textbf{0.70} (0.06)$ \\
        
        & CLUE &
        $0.38 (0.22)$ &
        $0.65 (0.02)$ &
        $0.41 (0.12)$ &
        $0.58 (0.02)$ &
        $0.54 (0.02)$ &
        $0.49 (0.01)$ &
        $0.07 (0.00)$ &
        $0.07 (0.00)$ \\
        
        & InfoShap &
        $0.41 (0.06)$ &
        $0.72 (0.04)$ &
        $0.52 (0.02)$ &
        $\textbf{0.88} (0.02)$ &
        $0.79 (0.01)$ &
        $\textbf{0.92} (0.01)$ &
        $0.59 (0.02)$ &
        $0.49 (0.05)$ \\ \midrule
    \multicolumn{10}{l}{\textbf{RMA}} \\ \midrule
         & VFA-IG &
        $0.57 (0.05)$ &
        $0.64 (0.03)$ &
        $0.79 (0.04)$ &
        $0.72 (0.07)$ &
        $0.83 (0.04)$ &
        $0.81 (0.08)$ &
        $0.50 (0.02)$ &
        $0.25 (0.03)$ \\
        
        & VFA-LRP &
        $0.57 (0.05)$ &
        $0.65 (0.04)$ &
        $0.76 (0.04)$ &
        $0.70 (0.08)$ &
        $0.82 (0.04)$ &
        $0.86 (0.05)$ &
        $0.49 (0.01)$ &
        $0.26 (0.03)$ \\
        
        & VFA-SHAP &
        $\textbf{0.83} (0.03)$ &
        $\textbf{0.92} (0.01)$ &
        $\textbf{0.89} (0.02)$ &
        $\textbf{0.87} (0.05)$ &
        $\textbf{0.95} (0.03)$ &
        $\textbf{0.94} (0.02)$ &
        $\textbf{0.75} (0.02)$ &
        $\textbf{0.44} (0.09)$ \\
        
        & CLUE &
        $0.34 (0.17)$ &
        $0.60 (0.02)$ &
        $0.27 (0.11)$ &
        $0.47 (0.01)$ &
        $0.52 (0.02)$ &
        $0.50 (0.01)$ &
        $0.07 (0.00)$ &
        $0.07 (0.00)$ \\
        
        & InfoShap &
        $0.38 (0.04)$ &
        $0.67 (0.04)$ &
        $0.41 (0.01)$ &
        $0.83 (0.03)$ &
        $0.79 (0.02)$ &
        $\textbf{0.94} (0.01)$ &
        $0.31 (0.01)$ &
        $0.26 (0.03)$ \\
        \bottomrule
    \end{tabular}%
}
\caption{Average local RRA and RMA over all test set instances for all considered uncertainty explainers and datasets (1-S: simple noise model and original train set, 50-C: complex noise model and artificially enlarged train set). Results are averaged across five folds, with standard deviations shown in brackets and best performances in bold. VFA consistently outperforms InfoSHAP and CLUE in all simple scenarios. VFA-SHAP also consistently performs best in all complex scenarios, while VFA-LRP and VFA-IG outperform CLUE in most cases and InfoSHAP in some complex scenarios.}\label{tab:localization}
\end{table}

As shown in \Cref{fig:cluecomparison}, CLUE assigns similar importance to all features. This effect is also present for InfoSHAP but is less pronounced. They are, therefore, less selective than VFA, potentially causing uncertainty features not to be detected for many instances. 

\begin{figure}[!hb]
    \centering
    \includegraphics[width=0.49\textwidth]{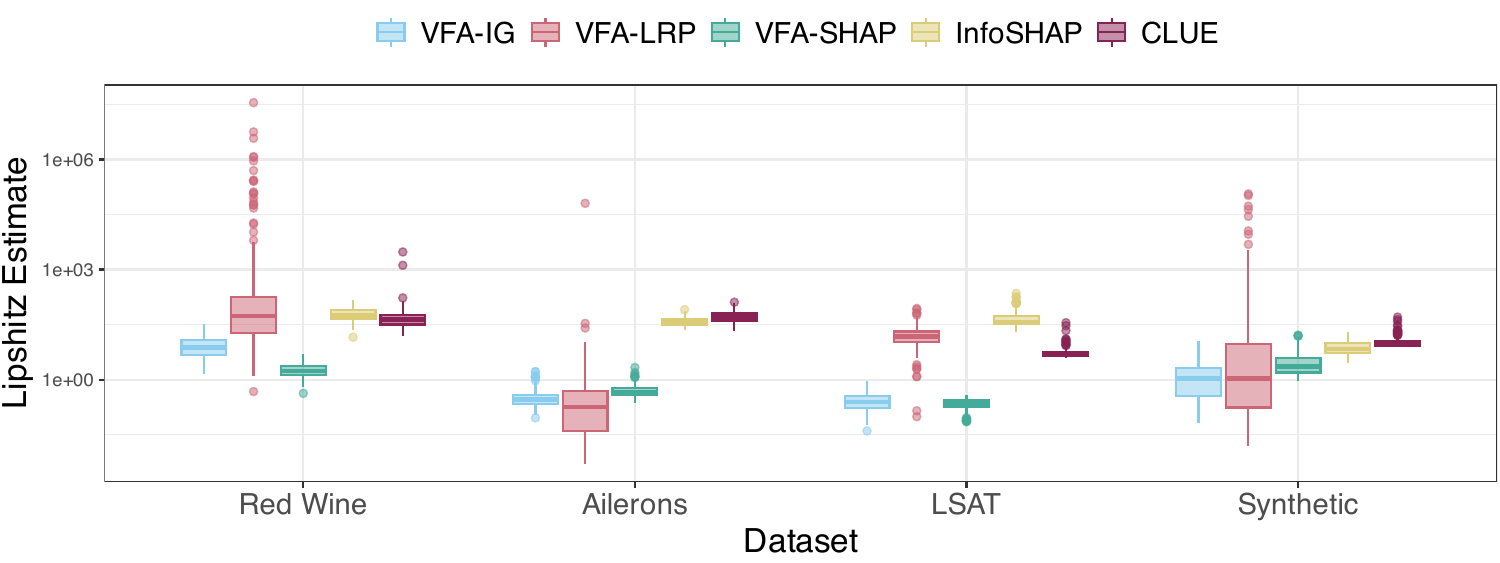}
   
  \caption{Local Lipschitz continuity estimates for 200 randomly chosen test set instances for all methods and datasets. Lower values indicate higher robustness. Having the lowest median Lipschitz estimates for most datasets, VFA-SHAP and VFA-IG are the generally more robust explainers.} 
\label{fig:lipschitz_rob}
\end{figure}

To analyze the robustness, we calculate distributions of local Lipschitz continuity estimates $\hat{L}(\vx_i)$ over 200 randomly chosen test set instances for each dataset and method (see \Cref{fig:lipschitz_rob}). According to the obtained Lipschitz estimates, VFA-SHAP and VFA-IG are generally more robust than InfoSHAP, CLUE, and VFA-LRP. The methods' individual ranking differs between datasets, suggesting that the choice of the most robust method is subject to the dataset.

When analyzing the faithfulness metric, we find that perturbation of the most important features faithfully reduces the correlation between uncertainties and residuals when VFA is used in the Ailerons and synthetic datasets (see \Cref{tab:perturbation}). However, VFA-LRP exhibits weaker faithfulness on the Ailerons dataset, demonstrating performance comparable to the baseline methods.

\begin{table}[!ht]
\centering
\scalebox{0.62}{

\begin{tabular}{@{}lcccc@{}}
\toprule
         & Red Wine & Ailerons  & Synthetic  \\ \midrule
VFA-IG   & -0.001±0.017 & -0.139±0.064 & \textbf{-0.167}±0.026  \\
VFA-LRP  & -0.004±0.020 & -0.083±0.042 & -0.154±0.028  \\
VFA-SHAP & 0.003±0.018  & \textbf{-0.171}±0.026 & -0.158±0.032  \\
InfoSHAP & -0.001±0.054 & -0.099±0.036 & -0.082±0.023 \\
CLUE     & -0.000±0.016 & -0.025±0.009 & -0.009±0.015  \\
\bottomrule
\end{tabular}}

\caption{Faithfulness of the uncertainty explanations: Change of Spearman correlation between uncertainties and squared residuals when most important features are perturbed. We expect faithful uncertainty explanations to induce a negative change. We exclude LSAT because we define perturbations only for continuous features. Results are the mean and standard deviation of 12 folds.}\label{tab:perturbation}
\end{table}

 On the small Red Wine dataset, the faithfulness of all methods near zero. In essence, the challenge of learning and explaining uncertainty is amplified in scenarios where data are scarce, leading to suboptimal faithfulness metrics.

\subsection{Benchmark on MNIST+U Image Data}
We evaluate the variance explainers on the MNIST+U dataset to understand their capability of dealing with higher-dimensional image data. We expect high attributions for pixels in the area of the uncertainty mask and, relative to that, low attribution on the mean mask.

\begin{figure}[!hbt]
    \centering
    \includegraphics[width=0.48\textwidth]{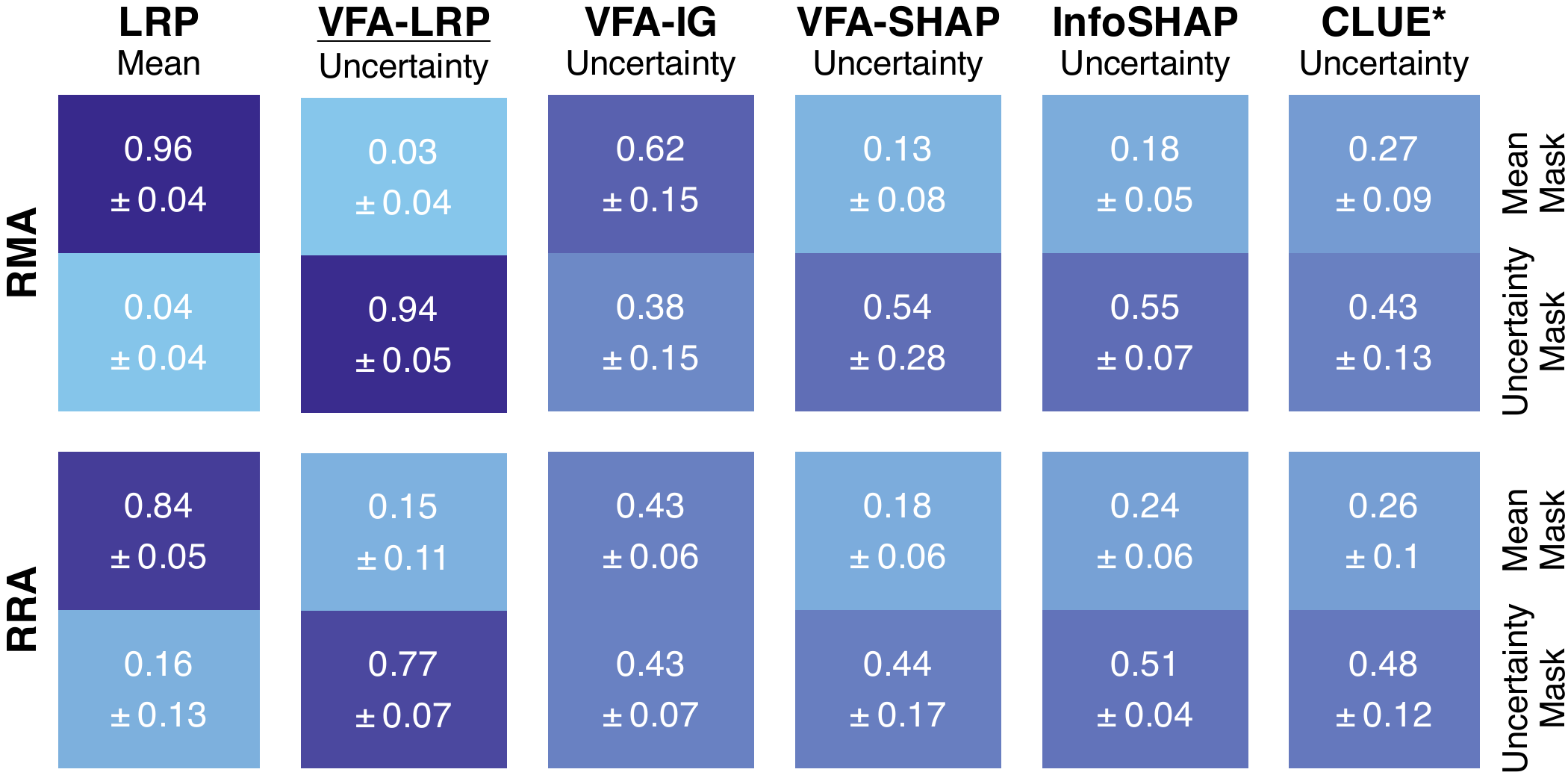}
    \caption{RMA and RRA for each uncertainty explainer and LRP mean explanations. We compare the attribution of pixels in the ground truth mask of the mean or the noise. We expect most of the relevance to be contained in the uncertainty mask. We show the mean and standard deviation over all samples in the test set. (*) Note: for CLUE, we only use 40\% of the test samples due to its high runtime.}
    \label{fig:mnist_heatmap}
\end{figure}

\begin{figure}[!b]
    \centering
    \includegraphics[width=0.49\textwidth, trim = 0 1.5cm 0 0, clip]{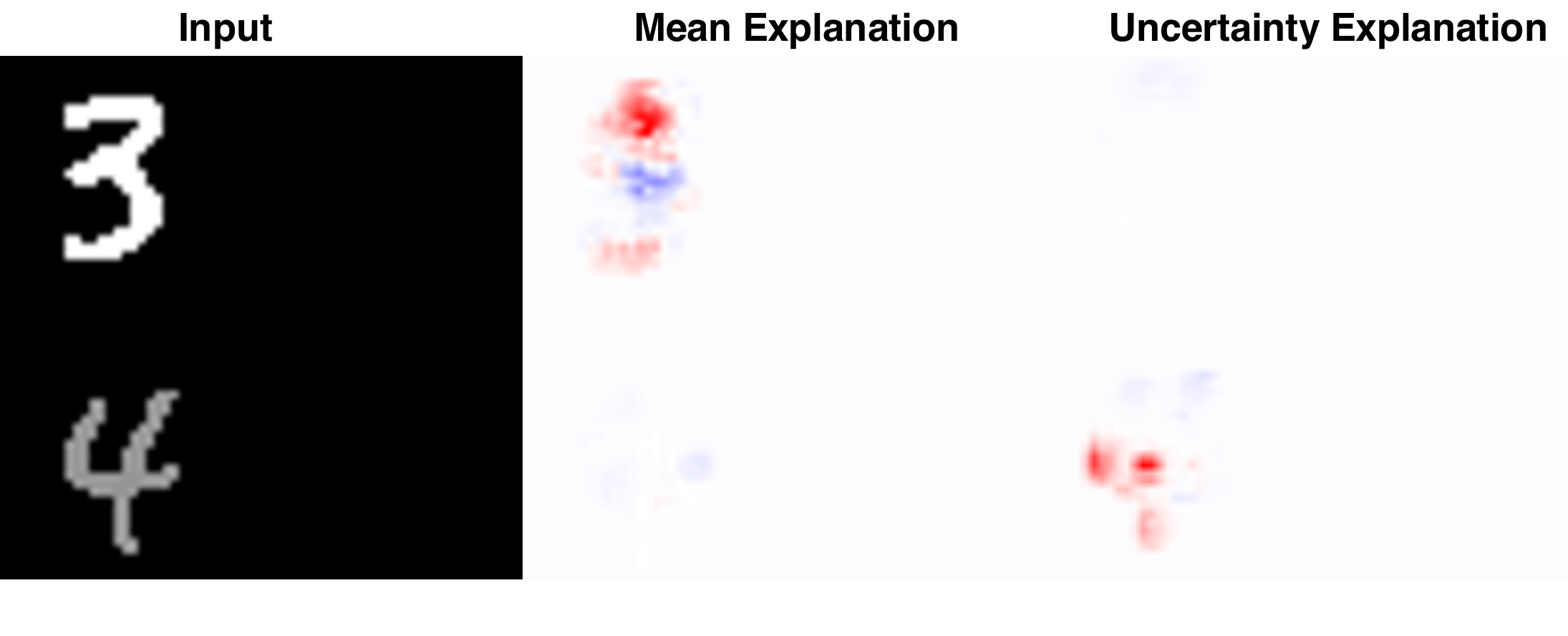}
    \caption{Mean and uncertainty explanations using LRP/VFA-LRP for a random sample from the MNIST+U test set. Both explanations focus on the digits relevant for mean and uncertainty, respectively.}
    \label{fig:mnist_lrp_example}
\end{figure}

All explanation methods, excluding VFA-IG, focus on the area of uncertainty mask (see \Cref{fig:mnist_heatmap}). VFA-LRP performs best, demonstrating the highest relevance attribution to the variance mask. This observation aligns with our expectation that most relevance should correspond to the variance, representing the primary source of uncertainty in the synthetic labels. We also see this in a randomly selected explanation example for VFA-LRP shown in \Cref{fig:mnist_lrp_example}.

While InfoSHAP and CLUE assign a considerable amount of relevance to the variance mask, they also have larger proportions of the variance explanation assigned to the mean mask.

VFA-IG performs poorly and assigns similar amounts of relevance to mean and variance masks, highlighting that the choice of the explanation method strongly influences the results. This aligns with findings that IG explanations on images focus on specific pixels rather than relevant patterns driving the prediction \citep{samek_explaining_2021}. This emphasizes the importance of selecting an appropriate explainer tailored to the specific characteristics of the data and uncertainty sources, which can be achieved in practice by considering faithfulness on a validation set.

\section{Discussion and Limitations}\label{sec:discussion_limitations}

We presented a straightforward strategy for explaining predictive aleatoric uncertainties, which requires minimal modifications to existing neural network regressors. We use neural networks with a Gaussian output distribution to estimate uncertainty and apply explanation methods to the variance output to explain uncertainty. In synthetic tabular experiments, the resulting explanations generally outperform alternative methods. As seen in the experiments with low uncertainty instances, we can also explain how features contribute to a model's certainty, which is relevant in high-risk applications. 
Since conventional evaluation metrics are not always directly applicable, we have introduced an evaluation protocol to assess uncertainty explainers. Parts of our evaluation depend on the knowledge of ground truth noise sources. This necessitated the incorporation of synthetic noise, which may deviate from the arbitrarily complex real-world noise patterns. We extend unsupervised explanation quality metrics for accuracy, faithfulness, and robustness to uncertainty attributions. In our benchmark, VFA compares favorably to CLUE and InfoSHAP. Using the MNIST+U image benchmark, we establish that the selection of the explanation method is a significant variable in generating high-fidelity uncertainty explanations. Generally, as we combine deep heteroscedastic regression with existing XAI methods, we inherit all the benefits and limitations of these methods, including computational complexity. 
A limitation of our analysis is its focus on aleatoric uncertainty. Approaches to model and explain epistemic uncertainty are equally crucial for safety-critical settings and should be used alongside our approach \citep{bley_explaining_2025}.
Future work might involve studying synergies in explaining point and uncertainty predictions. For example, in the context of explainable active learning \citep{ghai_explainable_2021}, a visualization of both explainability modes could be beneficial.

\section*{Ethical Statement}
There are no ethical issues.

\section*{Acknowledgments}
This work is supported by a BMWK grant (DAKI, 01MK21009E), a BMBF grant (act-i-ml, 01IS24078B), and a European Research Council grant (eXplAInProt, 101124385).
\newpage

\bibliographystyle{named}
\bibliography{references}

\end{document}